\documentclass[nohyperref]{article}

\usepackage{microtype}
\usepackage{graphicx}
\usepackage{subfigure}
\usepackage{booktabs} % for professional tables

\usepackage{hyperref}

\usepackage[accepted]{icml2022}

\usepackage{amsmath}
\usepackage{amssymb}
\usepackage{mathtools}
\usepackage{amsthm}
        \usepackage{multirow}

\usepackage[capitalize,noabbrev]{cleveref}
\newcommand{\ourFramework}{HyperInvariance}
\newcommand{\keypoint}[1]{\noindent\textbf{#1}\quad}

\newcommand{\cut}[1]{}

\theoremstyle{plain}
\newtheorem{theorem}{Theorem}[section]

\theoremstyle{definition}

\theoremstyle{remark}

\usepackage[textsize=tiny]{todonotes}

\icmltitlerunning{HyperInvariances: Amortizing Invariance Learning}

\begin{document}

\twocolumn[
\icmltitle{HyperInvariances: Amortizing Invariance Learning}

% It is OKAY to include author information, even for blind
% submissions: the style file will automatically remove it for you
% unless you've provided the [accepted] option to the icml2022
% package.

% List of affiliations: The first argument should be a (short)
% identifier you will use later to specify author affiliations
% Academic affiliations should list Department, University, City, Region, Country
% Industry affiliations should list Company, City, Region, Country

% You can specify symbols, otherwise they are numbered in order.
% Ideally, you should not use this facility. Affiliations will be numbered
% in order of appearance and this is the preferred way.
\icmlsetsymbol{equal}{*}

\begin{icmlauthorlist}
\icmlauthor{Ruchika Chavhan}{ed}
\icmlauthor{Henry Gouk}{ed}
\icmlauthor{Jan Stühmer}{sam}
\icmlauthor{Timothy Hospedales}{ed,sam}
% \icmlauthor{Firstname5 Lastname5}{yyy}
% \icmlauthor{Firstname6 Lastname6}{sch,yyy,comp}
% \icmlauthor{Firstname7 Lastname7}{comp}
% %\icmlauthor{}{sch}
% \icmlauthor{Firstname8 Lastname8}{sch}
% \icmlauthor{Firstname8 Lastname8}{yyy,comp}
%\icmlauthor{}{sch}
%\icmlauthor{}{sch}
\end{icmlauthorlist}

\icmlaffiliation{ed}{School of Informatics, University of Edinburgh}
\icmlaffiliation{sam}{Samsung AI Research, Cambridge}
% \icmlaffiliation{sch}{School of ZZZ, Institute of WWW, Location, C ountry}

\icmlcorrespondingauthor{Ruchika Chavhan}{r.chavhan@sms.ed.ac.uk}
% \icmlcorrespondingauthor{Firstname2 Lastname2}{first2.last2@www.uk}
% You may provide any keywords that you
% find helpful for describing your paper; these are used to populate
% the "keywords" metadata in the PDF but will not be shown in the document
\icmlkeywords{Machine Learning, ICML}

\vskip 0.3in
]

% this must go after the closing bracket ] following \twocolumn[ ...

% This command actually creates the footnote in the first column
% listing the affiliations and the copyright notice.
% The command takes one argument, which is text to display at the start of the footnote.
% The \icmlEqualContribution command is standard text for equal contribution.
% Remove it (just {}) if you do not need this facility.

%\printAffiliationsAndNotice{}  % leave blank if no need to mention equal contribution
\printAffiliationsAndNotice{\icmlEqualContribution} % otherwise use the standard text.

\begin{abstract}
Providing invariances in a given learning task conveys a key inductive bias that can lead to sample-efficient learning and good generalisation, if correctly specified. However, the ideal invariances for many problems of interest are often not known, which has led both to a body of engineering lore 
%about what often works in practice, 
as well as attempts to provide frameworks for  invariance learning. However, invariance learning is expensive and data intensive for popular neural architectures. We introduce the notion of amortizing invariance learning. In an up-front learning phase, we learn a low-dimensional manifold of feature extractors spanning invariance to different transformations using a hyper-network. Then, for any problem of interest, both model and invariance learning are rapid and efficient by fitting a low-dimensional invariance descriptor an output head. Empirically, this framework can identify appropriate invariances in different downstream tasks and lead to comparable or better test performance than conventional approaches. Our \ourFramework{} framework is also theoretically appealing as it enables generalisation-bounds that provide  an interesting new operating point in the trade-off between model fit and complexity. 
\end{abstract}

\section{Introduction}
Exploiting symmetries is crucial to the success of many intelligent systems. Convolutional neural networks most famously provide automatic translation equivariance, which improves learning for a large set of problems where this is an appropriate assumption. Symmetries can be seen as an example of inductive bias, which constrains the search space, or hypothesis class, of a learner. When well matched to the problem at hand, this leads to improvements in learning speed and generalisation and robustness to spurious correlations \cite{geirhos2020shortcut}. 

The importance of symmetries and invariances has led to research on architectures to enforce specific inductive biases', such as rotation and scale equivariance \cite{worrall2017harmonic,worrall2019deep}  (cf: translation). However, for many problems the ideal invariances are not known a-priori. This has led to a growing body of work on providing the ability to learn invariances, for example, by MAP learning \cite{benton2020invariances}, marginal-likelihood learning \cite{immer2022invariance} and meta-learning \cite{zhou2021metalearningSymmetries} (finding a common invariance that works well for a given problem family). Perhaps the most popular way to imbue general architectures with particular invariances  is using data augmentation. Optimising for the same output under a given transformation leads to invariance to that transformation \cite{benton2020invariances}. However, while flexible and simple to implement, this is extremely costly and data intensive. 

In a distinct body of work, research has identified a set of augmentations (and hence implicitly invariances) which are widely useful in representation learning for many popular tasks in computer vision  \cite{cubuk2019autoAug}.
%n. In supervised learning, this is exemplified by AutoAugment \cite{cubuk2019autoAug}. 
In the explosively growing field of self-supervised learning, this is exemplified by the suite of augmentations used by popular learners such as SimCLR \cite{chen2020simpleCLR}, which have been analysed in terms of the resulting invariances \cite{ericsson2021whyTransfer}. %Todo: Say some ore here.
However, while these augmentation suites are effective for many problems, they are not ideal for all problems  \cite{ericsson2021whyTransfer,xiao2021whatContrastive}. For example, while object recognition may prefer rotation invariance, dense prediction may prefer rotation equivariance. 
%This has in turn led various studies to attempt to develop augmentation learners that are efficient enough for problem-specific augmentation (invariance) learning \cite{raghu2021metaPT}. Nevertheless, the best of these algorithms are still extremely costly in both compute and data requirements.

In this paper we explore the idea of \emph{amortising} invariance learning. In an up-front step, we aim to train a feature extractor that compactly encodes a range of possible invariances. More specifically, we parameterize the feature extractor in terms of a hypernetwork \cite{ha2017hypernet}, which is conditioned on an invariance descriptor. This \ourFramework{} architecture thus defines a low-dimensional manifold of feature extractors defined by hypernetwork inputs, with points on the manifold corresponding to different invariance properties. In a task-specific step, we take this frozen (hyper) feature extractor and learn a new prediction head as well as the hypernetwork inputs, which correspond to choice of invariance for this task. In practice this framework means that new tasks can be solved -- including both model and invariance learning -- quickly, data-efficiently, and with few learnable parameters. Because invariances are selected out of a low-dimensional set,  they can be chosen very efficiently compared to existing approaches that train general neural architectures for invariances from scratch \cite{raghu2021metaPT,immer2022invariance,zhou2021metalearningSymmetries}. 

From a learning-theoretic point of view, \ourFramework{} is also appealing. Generalisation bounds guarantee testing error in terms of empirical risk (training error) plus model complexity. Standard frameworks provide different trade offs between these terms. Flexible deep models provide a good train fit, but poor model complexity, while shallow models provide a weaker train fit, but limited model complexity. \ourFramework{} provides an interesting new operating point. By squeezing a rich and relevant variety of feature extractors into a low-dimensional space of hyper-network inputs, we can provide an improved training fit with limited additional model complexity. 

\section{Related Work}
\keypoint{Invariance Learning} Invariances have been created by mean embeddings \cite{lyle2020benefits} and learned by MAP \cite{benton2020invariances}, marginal likelihood \cite{immer2022invariance}, or meta learning \cite{zhou2021metalearningSymmetries,cubuk2019autoAug,raghu2021metaPT} -- where gradients from the validation set are backpropagated to update the invariances or augmentation distributions. These mostly aim to learn task-specific symmetries, except \cite{zhou2021metalearningSymmetries}  who learn symmetries for a given family of tasks. All these approaches are highly data and compute intensive due to the substantial effort required to train such symmetries into general purpose architectures. Our \ourFramework{} framework amortises the cost of invariance learning so that it is quick and cheap to learn task-specific invariances downstream.

\keypoint{Invariances in Self Supervision} Self-supervised methods \cite{jing2021ssrlSurvey,ericsson2022ssrlSurvey} often rely on contrastive augmentations \cite{chen2020simpleCLR}. Their success has been  interpreted as engendering invariances \cite{ericsson2021whyTransfer,wang2020alignUniform,purushwalkam2020demystifying} through these augmentations, which in turn provide good inductive bias for downstream tasks. While self-supervision sometimes aspires to providing a single general purpose feature suited for all tasks in the guise of foundation models \cite{bommasani2021foundation}, studies have shown that different augmentations (invariances) are suited for different downstream tasks, with no single feature being optimal for all tasks. This leads to the tedious need to produce and combine an ensemble of features \cite{xiao2021whatContrastive,ericsson2021whyTransfer}, or task-specific self-supervised pre-training \cite{raghu2021metaPT}, which is extremely costly. Our \ourFramework{} breaths new life into the notion of general purpose features by defining a parametric feature extractor that spans an easily accessible range of invariances.

%tasks that are learned simultaneously in a multi-task learning setup. 
%We consider a set of $\mathcal{T} = \{\mathcal{T}_{\text{train}}, \mathcal{T}_{\text{test}} \}$ with different invariance requirements.
\section{Methodology}

\keypoint{Pre-train}
We begin by considering a set of pre-training tasks $\mathcal{T}_{\text{train}}$ with dissimilar invariance requirements. Let the dataset corresponding to task $t$ be $\mathcal{D}^t = \{ x^t_i, y^t_i\}_{i=1}^{n_t}$, where $n_t$ is the number of samples per-task. Let $f_\theta$ and $\Phi_{\text{train}} = \{\phi^t\}_{t \in \mathcal{T}_{\text{train}}}$ be the shared feature extractor and task-specific decoder weights for training tasks respectively. Predictions for a task $t$ is given by $\hat{y}^t = \langle \phi^t, f_\theta(x^t) \rangle$.

During pre-training the invariance descriptor  $i^t$ for each task is assumed observed. This is a  vector $i\in[0,1]^K$ over $K$ possible transformations, where $i_k=1$ and $i_k=0$ indicate invariance and sensitivity to the $k$th factor respectively. 

We employ hypernetworks to generate weights for the given encoder given the desired invariance descriptor $i^t$. 
%Usually, hypernetworks are designed as two layer networks that generate weights for a larger neural networks given an embedding vector. 
In this setup, we use a hypernetwork which is shared between all training tasks to generate weights for the encoder to satisfy task-specific invariance requirements.

% Write about hyper-networks  and how hypernetwork is shared between tasks
%We denote the 2-layer hypernetwork by $h$ parametrized by $W = \{ w_1, w_2, b_1, b_2\}$. The hypernetwork $h$ generates weights for the encoder given a invariance hyper-parameter for task $t$ denoted by $i_t$ as $h(i) = w_2^T(\sigma(w_1^T i) + b_1) + b_2$, where $\sigma$ is a non-linear activation function. 
The hypernetwork $h$, parametrized by $W$ generates weights for the encoder given a invariance hyper-parameter $i_t$ for task $t$  as $\theta=h_W(i_t)$. Thus, predictions for task $t$ are 
\begin{equation}
    \hat{y}^t = \langle \phi^t, f_{h(i_t)}(x^t) \rangle\label{eq:mtl}
\end{equation}
%Idk if i should use the words "meta-training" 
In the pre-training stage, the hypernetwork and the corresponding task-specific parameters are updated to optimize the loss of the training tasks.
\begin{equation}
    W^\star, \Phi_{\text{train}}^\star = \arg \min_{W, \Phi_{\text{train}}} \frac{1}{|\mathcal{T}_{\text{train}}|}  \sum_{t \in \mathcal{T}_{\text{train}}} \frac{1}{n_t} \sum_{j=1}^{n_t}  \mathcal{L}(\hat{y}^t_j, y^t_j)\label{eq:mtl2}
\end{equation}

\keypoint{Downstream} We next consider a set of downstream tasks $\mathcal{T}_{\text{test}}$, that can be solved optimally with different invariance requirements which are unknown to us. We denote the training data available for a downstream task $t^\prime \in \mathcal{T}_{\text{test}}$ as $\mathcal{D}^{t^\prime} = \{x^{t^\prime}_i, y^{t^\prime}_i \}_{i=1}^{n_{t^\prime}}$. In the downstream task training stage, we employ the optimal hypernetwork learned from the pre-training stage to make predictions for the test-tasks. For a downstream task $t^\prime$, let $ \hat{y}^{t^\prime} = \langle \phi_t, f_{h^\star(i_t)}(x^{t^\prime}) \rangle$ be the predicted output given a invariance hyper-parameter $i_{t}$. Subsequently, we evaluate the optimal invariance hyper-parameters and prediction heads $ \Phi_{\text{test}}$ by minimzing the task-specific loss of the training set,
\begin{equation}
    i^\star_t, \Phi_{\text{test}}^\star = \arg \min_{i_t, \Phi_{\text{test}}} \frac{1}{|\mathcal{T}_{\text{test}}|}   \sum_{t^\prime \in \mathcal{T}_{\text{test}}} \frac{1}{n_{t^\prime}} \sum_{j=1}^{n_{t^\prime}}  \mathcal{L}(\hat{y}^{t^\prime}_j, y^{t^\prime}_j).
\label{eq:metatest}
\end{equation}

\keypoint{Continuous vs Discrete Invariances Estimates}
This setup provides the option of (i) using continuous invariance estimates $i^*_t$ and $\Phi^*_t$ directly, or (ii) discretizing invariance estimates, $\tilde{i}^\ast = \text{round}(i^*_t)$, before re-learning $\Phi$. The latter may provide an advantage if $i^*_t$ can be stuck in a local minima, and also provides a learning theoretic advantage.

\keypoint{Theoretical Analysis}
We provide some theoretical analysis into the potential of our approach to overfit when applied to novel downstream tasks not seen during pre-training in the theorem below.
\begin{theorem}
\label{thm:main}
For 1-Lipschitz loss function, $\mathcal{L}$, ff for all $\phi$ we have that $\|\phi\| \leq B$ and $\|f_\phi(x)\| \leq X$, the following holds with probability $1 - \delta$
\begin{equation*}
    \mathbb{E}_{x^t, y^t} [\mathcal{L}(\hat{y}^t, y^t)] \leq \frac{1}{n_t} \sum_{j=1}^{n_t} \mathcal{L}(\hat{y}^t_j, y^t_j) + \frac{2XB}{\sqrt{n_t}} + 3\sqrt{\frac{\textup{ln}(|I|/\delta)}{2n_t}},
\end{equation*}
where $I = {\{0, 1\}}^d$ is the space of possible invariance hyperparameters.
\end{theorem}
The proof is in the appendix. This theorem demonstrates that the overfitting behaviour of our approach when applied to novel tasks scales similarly to conventional linear models (i.e., $\frac{2XB}{\sqrt{n}}$), but due to the more flexible feature extractor we have a potential to obtain a much better fit on the training data, and therefore a better guarantee on test error. In contrast, formal analysis of the overfitting behaviour of deep neural networks cannot obtain as tight bounds, instead depending exponentially on the depth of the network due to the product of norms of the weights in each layer \citep{bartlett2017spectrally,golowich2018size,long2019generalization}.

\section{Experiments and Results}
\label{expts-section}
% We want to investigate if the HyperInvariance model is better at generating encoder weights that capture the properties optimal for the downstream task.
% Dataset, tasks and invariance descriptors

We conduct synthetic and real-world learning experiments.

\cut{\textbf{Hypernetwork architecture:} We use a 2-layer hypernet $h$ with $W = \{ w_1, w_2, b_1, b_2\}$ so $h_W(i) = w_2^T(\sigma(w_1^T i) + b_1) + b_2$, where $\sigma$ is a non-linear activation function. }
%We denote the 2-layer hypernetwork by $h$ parametrized by $W = \{ w_1, w_2, b_1, b_2\}$. The hypernetwork $h$ generates weights for the encoder given a invariance hyper-parameter for task $t$ denoted by $i_t$ as $h(i) = w_2^T(\sigma(w_1^T i) + b_1) + b_2$, where $\sigma$ is a non-linear activation function. 

%, where $1$ stands for complete invariance and $0$ stands for sensitivity. Thus, when we perform experiments with only rotation/color invariance, we pass $i=1$ and $i=0$ for digit prediction and rotation/color prediction respectively. Thereafter, for experiments with $k$ invariances $i$ becomes a binary vector with $k$ bits, each bit specifying if the task is invariant or sensitive for that particular property. 

%We focus on two types of training paradigms for our experiments namely synthetic experiments on multi-task learning and real-world experiments on contrastive learning. 

\textbf{Synthetic experiments:} Firstly, we perform a set of simple experiments consisting of hand-designed tasks with distinct invariance requirements trained jointly in a multi-task learning setup. \emph{Pre-Train:} We consider  color and rotation invariance on a Colored-Rotated MNIST dataset, which is generated by randomly coloring one of the RGB channels of grayscale Rotated MNIST images. In our experiments, Rotated MNIST is generated by rotating MNIST images by randomly selecting an angle from the set $\{-90, -60, -30, 0, 30, 60, 90\}$. We consider three tasks: (i) digit prediction, (ii) rotation prediction and (iii) color prediction. In this setup, digit prediction is ideally color and rotation invariant, while rotation angle prediction is rotation-sensitive and color-invariant and color prediction is rotation-invariant and color-sensitive. Following the convention of assigning invariance descriptors, we assign the invariance hyper-parameters to be $i = [1,1]$, $i=[1,0]$ and $i=[0,1]$ for digit, color, and rotation prediction respectively. We use Augerino \cite{benton2020invariances} augmentations to teach the invariances corresponding to each hyper-parameter setting. With these tasks, we learn a hypernet derived feature encoder and distinct readout heads (Eqs.~\ref{eq:mtl} and \ref{eq:mtl2}). \emph{Downstream:} As a downstream task, we consider digit prediction and rotation prediction on a new domain of Colored-Rotated KMNIST, generated in the same way as the pre-training dataset to examine if the HyperInvariance model can learn appropriate invariances with limited data. Here the (hyper) feature extractor is frozen, and a new readout head and hyperparameters (Eq~\ref{eq:metatest}) are learned. \emph{Competitor:} For a baseline we do multi-task learning (MTL) among source tasks with a conventional shared feature extractor, and learn a new head for the downstream task.

\begin{table}[]
    \centering
    \resizebox{\columnwidth}{!}{
    \begin{tabular}{c|cccc|cccc}
    \hline
        & \multicolumn{4}{c|}{Digit Prediction} & \multicolumn{4}{c}{Rotation Prediction} \\
        \hline
        $N$ & $i_{\star}$ & 
        $A_{I_\star}$ & $A_{i_\star}$ & $A_{\text{MTL}}$  & $i_{\star}$ & 
        $A_{I_\star}$ & $A_{i_\star}$ & $A_{\text{MTL}}$\\
        \hline
        10 & [61, 65] & 27.5 & \textbf{33.3} & 22.8 & [35, 75] & 52.9 & \textbf{55.8} & 45.7\\
        20 & [60, 74] & 41.9 & \textbf{42.1} & 35.2 & [ 6, 88] &  72.1 &  \textbf{74.4} & 70.0\\
        50 & [63, 80] & 47.9 & \textbf{49.2} & 46.7  & [2, 93] & 77.8 & \textbf{81.1} & 74.1  \\
        100 & [65, 86] &  51.3 & \textbf{52.5} & 47.0  & [ 0 , 86] & 79.7 & \textbf{84.4} & 76.3 \\
        200 & [72, 91] & 51.5 & \textbf{54.2} & 48.0 & [ 0, 90] & 85.3  & \textbf{86.9} & 83.0\\
        \hline
    \end{tabular}}
    \caption{Digit and Rotation prediction on ColoredRotated-KMNIST. $i^*$: estimated invariance strength (\%) to (rotation, color). $A_{I_*}, A_{i_*}$: HyperInvariance accuracy (\%) with binarized and continous invariance respectively. $A_{MTL}$ Multi-task baseline accuracy.}
    \label{tab:synthetic-exps}
\end{table}
% Describing SimCLR experiments
% Describe dorsal, ventral, default
% Resnet 18 and hypernetwork

% Three tables 
% Three plots
\begin{table}[]
    \centering
    \resizebox{\columnwidth}{!}{
    \begin{tabular}{|c|ccc|ccc|}
    \hline
        Task & \multicolumn{3}{c|}{HyperSimCLR} & \multicolumn{3}{c|}{Competitors} \\
             & $i^{\star}$ & $A_{I_\star}$ &$A_{i_\star}$ & SimCLR & Ventral-SimCLR & Dorsal-SimCLR \\
        \hline
        300W & [46, 59] & 74.0 & \textbf{84.0} & 45.0 & 51.0 & 82.0 %DPs
        \\
        CIFAR10 & [99, 96] & 76.9 & 78.4 & \textbf{78.6} & 68.4 & 50.5\\
        \hline
    \end{tabular}}
    \caption{Quantitative results for two downstream tasks after SimCLR-ResNet18 training on STL10: CIFAR recognition (accuracy, \%) and 300W landmark detection ($R^2$, \%). Our HyperSimCLR outperforms regular SimCLR \cite{chen2020simpleCLR} on 300W, and Dorsal/Ventral SimCLRs \cite{ericsson2021whyTransfer} on both.}
    \label{tab:my_label}
\end{table}

\begin{figure*}
    \centering
    \includegraphics[width=0.325\linewidth]{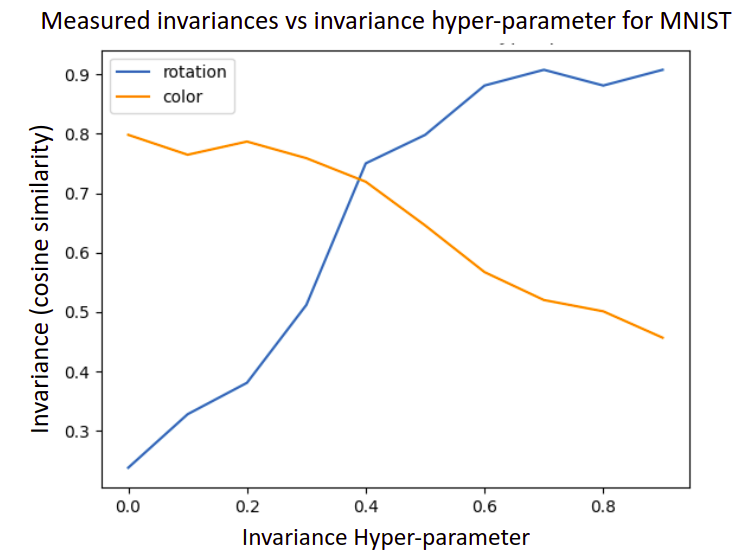}
    \includegraphics[width=0.35\linewidth]{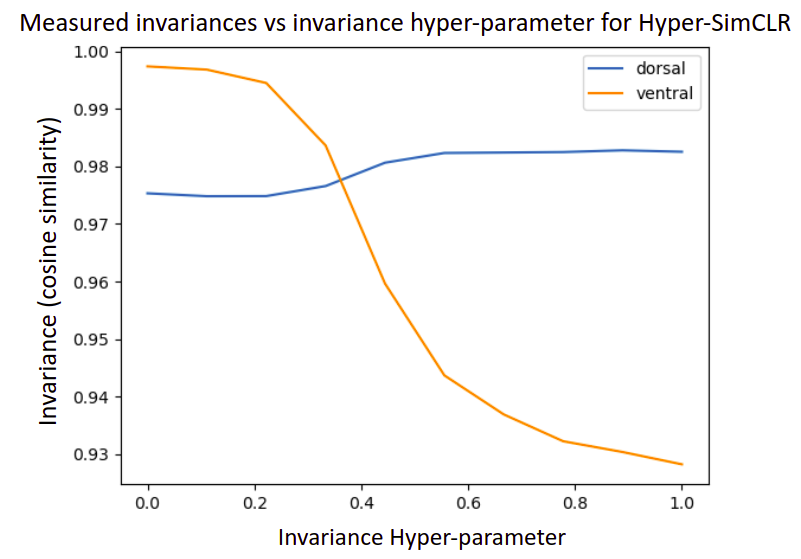}
    \includegraphics[width=0.31\linewidth]{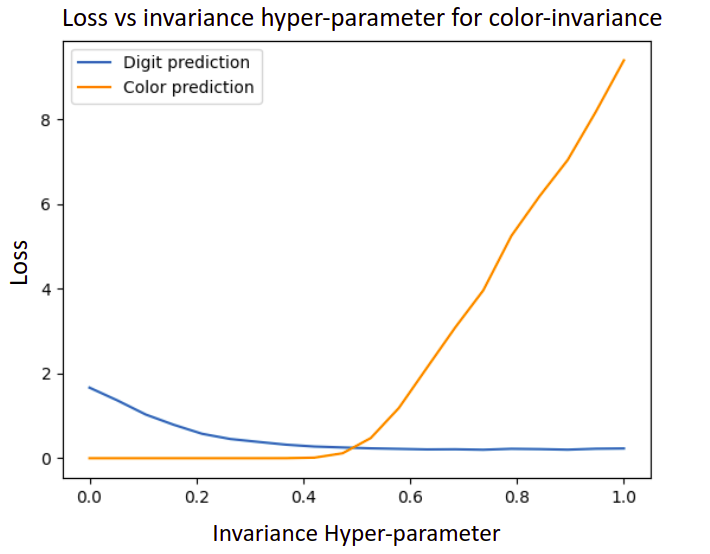}
    \vspace{-0.2cm}
%    \caption{Loss for digit prediction (blue) and color prediction (orange) for color invariance experiments}
%    \label{fig:acc-vs-inv-r}
% \end{figure}
% \begin{figure}
%     \centering
%    \caption{Measured invariance vs hypernetwork input for rotation and color invariance experiments. }
%    \label{fig:cos-vs-inv-c}
% \end{figure}
% \begin{figure}
%     \centering
%    \caption{Measured invariance vs hypernetwork input for SimCLR experiments. \textcolor{red}{Don't know why the variation in dorsal model is low and invariances are still high}}
    %\label{fig:cos-vs-inv-c}
    \caption{Left: Measured invariances (to rotation and color) of hypernet features as a function of invariance hyperparameter - MNIST experiment. Middle: Measured invariance (to dorsal and ventral transforms) of HyperSimCLR-ResNet18 feature extractor as a function of  invariance hyperparameter - STL10. Right: Train loss for digit and rotation prediction vs invariance hyperparameter - STL10/HyperSimCLR.} \label{fig}
    \vspace{-2pt}
\end{figure*}

\textbf{Contrastive Learning experiment:} We next evaluate our \ourFramework{} framework with a real-world contrastive learning experiment, taking SimCLR \cite{chen2020simpleCLR}  as a representative state of the art learner  to build upon. To define a set of invariances of interest, we borrow  from \citet{ericsson2021whyTransfer}
who extract two subsets of SimCLR augmentations denoted  \textit{dorsal} and \textit{ventral}. % and \textit{default} corresponding to appearance-based, spatial, and standard array of augmentations (both). 
\citet{ericsson2021whyTransfer}  showed that contrastive models trained to maximize similarity between images and their dorsal/ventral augmented counterparts learn representations invariant to corresponding transformations. \emph{Pre-train: }We instantiate our framework with SimCLR (hereafter referred to as HyperSimCLR) and a ResNet18 architecture and train on STL-10 \cite{coates2011stl10}. During training, we assign invariance hyper-parameters as $i = [1,1]$, $i=[1,0]$ and $i=[0,1]$ for default, ventral and dorsal augmentations respectively. For every $i$, the hypernetwork is trained to optimize the contrastive loss for an image and the counterpart with augmentation corresponding to $i$. \emph{Downstream} Given the learned frozen HyperSimCLR-ResNet18 feature, we train linear readout and hyperparameters for new tasks. For recognition we evaluate CIFAR10 classification, and for regression, we evaluate 300W facial interest point detection \cite{sagonas2016_300facesWild}.

\subsection{Results}
With the experiment setup above, we aim to answer the following three research questions: 

\keypoint{Does the HyperNetwork learn to generate invariant features?} To examine if the hypernetwork has learned features that are provide a specified invariance, we measure invariance by evaluating the cosine similarity between images and their augmented versions when providing different hypernetwork inputs. Figure \ref{fig}(left) reports the invariance 
%(cosine sim between digit and transformed counterpart)  %% said it already just abve ^
to rotation and color in the synthetic multi-task experiment. At each x-coordinate we pass $[i, 1-i]$ to the hypernetwork to interpolate between rotation and color invariance. The trend shows that invariances broadly vary between minimum and maximum as a function of hyperparameter $i$. Figure \ref{fig}(middle) performs the corresponding evaluation for dorsal and ventral augmentations in the SimCLR-ResNet18 experiment. Again we see that measured invariance is a near monotonic function of the hyperparameter. These results show that the hypernet can synthesize features with a desired invariance for both shallow and deep CNNs, and that both supervised multi-task and self-supervised pre-training can be used. 

\keypoint{Can the preferred invariance of a downstream task be identified?} Given our pre-trained hypernet (on MNIST and STL10 respectively) we study downstream tasks (KMNIST and CIFAR, 300W respectively) by learning a linear readout and invariance hyperparameters. First, for the synthetic experiment, we solve downstream KMNIST with a variety of invariance parameters, interpolating along [$i,1$] (color invariance). We report the training loss for digit prediction and color prediction.
%\doublecheck{$i$, and report the training loss for a single-invariance learning experiment. 
%Here, we present the loss for color-invariance experiments for both digit and color prediction. 
The results in Figure \ref{fig}(right) show that  loss is a clean monotonic function of the corresponding invariance: Digit prediction is best with maximum color invariance; color prediction is obviously best with minimum color invariance. Both curves are quite smooth, so we expect optimising loss wrt the invariance parameter (x-axis) will discover a good invariance parameter. 

Next we check if we can optimise downstream tasks wrt preferred invariance parameter (Eq.~\ref{eq:metatest}). The results for synthetic digit prediction and color prediction are shown in Table \ref{tab:synthetic-exps} for $N$ training samples per class. Inspecting Table~\ref{tab:synthetic-exps} for digit prediction, we see that invariances are tending towards $[1,1]$ (and always round to $[1,1]$) correctly identifying that digit prediction should be both color and rotation invariant. For rotation prediction in Table \ref{tab:synthetic-exps},  invariances are tending towards $[0,1]$ (and always round to $[0,1]$) confirming that color prediction should be only rotation invariant.

%the variation of measured invariances for digit prediction in synthetic experiments with respect to the invariance hyper-parameter passed as an input to the hypernetwork. As expected, the cosine similarity between features of original image and augmented image is low/high when the invariance descriptor for sensitivity/invariance is passed through the hypernetwork. Subsequently, we also measure invariances for HyperSimCLR experiments for the CIFAR10 dataset, augmented with only dorsal augmentations and pass $[i, 1 - i]$ to the hypernetwork for $i$ ranging between $0$ and $1$. A consistent pattern is observed where invariance descriptor specifying invariance to dorsal augmentations $i=[0,1]$ results in high cosine similarity. 

\textbf{Quantitative results:} For the synthetic experiment results in  Table \ref{tab:synthetic-exps} shows that (i) HyperInvariance outperforms the multi-task baseline in accuracy, especially at low samples per class -- demonstrating the benefit of the correctly discovered invariance as an inductive bias. (ii) Discretizing the invariances  usually performs only slightly worse ($A_{I_*}$ vs $A_{i_*}$), while providing theoretical benefits.

For the HyperSimCLR experiment in Table~\ref{tab:my_label}, we can see that for CIFAR-10: (i) Our HyperSimCLR selects both dorsal and ventral invariance (ii) it performs similarly to regular SimCLR baseline and both outperform the dorsal- and ventral-alone baselines. Meanwhile for 300W: (i) HyperSimCLR clearly outperforms regular SimCLR thanks to the ability to tune invariances to a moderate amount with a small preference for Dorsal invariances, and even slightly outperforms DorsalSimCLR which was previously best at this task. Overall, this solves the problem in \cite{ericsson2021whyTransfer} where specific choice of dorsal, ventral, or regular SimCLR had to be made on a per-problem basis. A single HyperSimCLR feature can support different tasks through amortised invariance learning.

% \vspace{-0.2cm}
\subsection{Discussion} We have shown that multiple invariances can be pre-learned by a single (hyper) feature extractor. This enables downstream tasks to easily select a useful invariance. This provides an interesting new avenue of study for general purpose features suited for diverse downstream tasks. 
\section*{Acknowledgements}
R.C. was supported by Samsung AI Research, Cambridge.
% \textbf{Do not} include acknowledgements in the initial version of
% the paper submitted for blind review.

% If a paper is accepted, the final camera-ready version can (and
% probably should) include acknowledgements. In this case, please
% place such acknowledgements in an unnumbered section at the
% end of the paper. Typically, this will include thanks to reviewers
% who gave useful comments, to colleagues who contributed to the ideas,
% and to funding agencies and corporate sponsors that provided financial
% support.

% \nocite{langley00}

\bibliography{example_paper}
\bibliographystyle{icml2022}

%%%%%%%%%%%%%%%%%%%%%%%%%%%%%%%%%%%%%%%%%%%%%%%%%%%%%%%%%%%%%%%%%%%%%%%%%%%%%%%
%%%%%%%%%%%%%%%%%%%%%%%%%%%%%%%%%%%%%%%%%%%%%%%%%%%%%%%%%%%%%%%%%%%%%%%%%%%%%%%
% APPENDIX
%%%%%%%%%%%%%%%%%%%%%%%%%%%%%%%%%%%%%%%%%%%%%%%%%%%%%%%%%%%%%%%%%%%%%%%%%%%%%%%
%%%%%%%%%%%%%%%%%%%%%%%%%%%%%%%%%%%%%%%%%%%%%%%%%%%%%%%%%%%%%%%%%%%%%%%%%%%%%%%
\newpage
\appendix
\onecolumn
\section{Additional Discussion On Theory}
We can contrast the result in Theorem~\ref{thm:main} with the corresponding result for standard approaches. 

The corresponding result for the popular protocol of linear readout from a fixed feature would reduce the third term in the right-hand side of the bound by eliminating the dependence on $|I|$. However, it would also worsen the empirical risk (first term, RHS). This is illustrated in Table~\ref{tab:appendix-exps} where we also show the train performance for HyperInvariance and the MTL baseline -- HyperInvariance provides a clearly better training fit. 

Meanwhile, the corresponding result for the other popular protocol of fine-tuning the whole feature extractor would improve the first empirical risk term in the bound, but introduce an exploding complexity term in the bound. In particular, the second term of the bound would be proportional to a product of norms of the weight matrices in each layer, thus scaling exponentially with the depth of the network. See, e.g., \citet{golowich2018size} for a demonstration of this in the framework of Rademacher complexity-based analysis.

This is why we describe HyperInvariance as providing an interesting new theoretical operating point between two popular protocols.

\keypoint{Connection to Empirical Results} 
With regard to the discretization operation, our current experiments report binary discretization of the continuously estimated invariance parameter, which works well in the synthetic experiments and CIFAR-10. We remark that our framework and Theorem are compatible with any quantization strength, for example ternary or higher. This may provide a better tradeoff between empirical performance + theoretical guarantees for benchmarks like 300W, where the strong results for our continuous model ($A_{i_*}$) in Tab~\ref{tab:my_label} suggest that a moderate amount of invariance is preferred. 

\subsection{Proof of Theorem \ref{thm:main}}
\begin{proof}
There is a one-to-one mapping between elements of the finite set of invariance hyperparameters and potential feature extractors, implying that there is also a finite number, $|I|$, of potential feature extractors for a novel task. For a linear model coupled with a pre-selected feature extractor, one can apply the standard generalisation bound for linear models based on (empirical) Rademacher complexity \citep{bartlett2002rademacher} to obtain, with probability $1-\delta$,
\begin{equation*}
    \mathbb{E}_{x^t, y^t} [\mathcal{L}(\hat{y}^t, y^t)] \leq \frac{1}{n_t} \sum_{j=1}^{n_t} \mathcal{L}(\hat{y}^t_j, y^t_j) + \frac{2XB}{\sqrt{n_t}} + 3\sqrt{\frac{\textup{ln}(1/\delta)}{2n_t}}.
\end{equation*}
Our result follows from using the union bound to optimise over the choice of feature extractor.
\end{proof}

\section{Additional Details}

% Things to write about: convolution kernel size, hypernetwork architecture, feature dimension, learning rates, schedulers
\subsection{Synthetic experiments} 

\textbf{Hypernetwork:} For the multi-task learning experiments, we fix $f_\theta$ to be a single convolution layer with 16 filters of kernel size $5 \times 5$. This convolution is performed with a stride of 2 and is followed by batch normalization and ReLU before passing the features to the task-specific heads. The hypernetwork is designed as a two-layer network generates weights for the convolution layer given binary invariance descriptors . We design the hypernetwork such that $w_1 \in \mathbb{R}^{k \times d_\text{h}}$, $b_1 \in \mathbb{R}^{d_h}$, $w_2 \in \mathbb{R}^{d_\text{h} \times d_\text{out}}$ and $b_2 \in \mathbb{R}^{d_{\text{out}}}$. Here, we choose $d_\text{h} = 40$ and subsequently $d_\text{out} = 1200$ which is equal to the total number of weights of the convolution layer. For colored and rotated MNIST images, the output of the hypernetwork is further resized as $3 \times 16 \times 5 \times 5$. 

% \textbf{Datasets:} As mentioned in Section \ref{expts-section} in tha main paper, we perform all pre-trainig experiments on a Colored and Rotated version of the MNIST dataset. We convert rotated grayscale images into colored images by randomly filling one of the RGB channels and train the hypernetwork to predict this color channel and the rotation angle. 

\textbf{Training Details:} For digit prediction, color predictions, and rotation prediction the output cardinality of the task-specific prediction heads is 10, 3, and 7 respectively. During the pre-training stage, the hypernetwork and the task-specific weights are trained for 200 epochs with the Adam optimizer with a learning rate of $5e^{-4}$ along with a cosine annealing learning rate scheduler. During the downstream task learning stage, the invariance hyper-parameters and task-specific decoders are trained with the Adam optimizer with the same learning rate as above.

\begin{table*}[]
    \centering
    \begin{tabular}{c|ccccc|ccccc}
    \hline
        & \multicolumn{5}{c|}{Digit Prediction} & \multicolumn{5}{c}{Rotation Prediction} \\
        \hline
        $N$ & $i_{\star}$ & 
        $A^{\text{train}}_{i^\star}$ & $A_{i_\star}$ & $A_{\text{MTL}}^\text{train}$& $A_{\text{MTL}}$  & $i_{\star}$ & 
        $A^{\text{train}}_{i_\star}$ & $A_{i_\star}$ & $A_{\text{MTL}}^\text{train}$ & $A_{\text{MTL}}$\\
        \hline
        10 & [61, 65] & 98.2 & 33.3 & 59.8 & 22.8 & [35, 75] & 100.0 & 55.8 & 55.8 & 45.7\\
        20 & [60, 74] & 98.1 & 42.1 & 71.6 & 35.2 & [ 6, 88] &  100.0 & 74.3 &  74.4 & 70.0\\
        50 & [63, 80] & 95.7 & 49.2 & 73.2& 46.7  & [2, 93]  &97.1 & 81.1 & 81.1 & 74.1  \\
        100 & [65, 86] &  87.7 & 52.5 & 64.5 & 47.0  & [ 0 , 86] & 84.4 & 79.7 & 84.4 & 76.3 \\
        200 & [72, 91] & 82.3 & 54.2 & 72.9& 48.0 & [ 0, 90] & 92.5 & 86.9 & 86.9 & 83.0\\
        \hline
    \end{tabular}
    \caption{Digit and Rotation prediction on ColoredRotated-KMNIST. $i^*$: estimated \% invariance to (rotation, color). $A^{\text{train}}_{i^\star}$: HyperInvariance train accuracy with continuous invariance, $A_{i_*}$: HyperInvariance test accuracy with continuous invariance. $A_{\text{MTL}}^\text{train}$: Multi-task baseline train accuracy. $A_{MTL}$ Multi-task baseline test accuracy.}
    \label{tab:appendix-exps}
\end{table*}

\subsection{Hyper-SimCLR}

\textbf{Hypernetwork:} For contrastive learning experiments, the hypernetwork is trained to generate the parameters of the convolution layers of the Resnet18 model. Unlike the synthetic experiments, the hypernetwork generates kernels for multiple convolution layers of the Resnet18 model.  For SimCLR experiments with two types of invariances ($k=2$), the hypernetwork is designed as a two layer-network with and $d_\text{h} = 64$ such that $w_1 \in \mathbb{R}^{2 \times 64}$ and $b_1 \in \mathbb{R}^{64}$. However, to generate a convolution layer with parameters denoted by $\{\theta_l\}_{l=1}^{18}$ of the Resnet18 architecture, we learn a separate set of $\{w_2^l, b_2^l\}_{l=1}^{18}$ such that $\theta_l = {w_2^l}^T(\sigma(w_1^T i) + b_1) + b_2^l$. We direct the readers to \cite{ha2017hypernet} for more details on the hypernetwork architecture that generates all convolution kernels for Resnet18. 

\textbf{Augmentations:} In this section of the appendix, we provide details about the augmentation policies used to train Hyper-SimCLR \cite{ericsson2021whyTransfer}. Ventral augmentation is a set of spatial transformations including random resized crop and random horizonal flip. Dorsal augmentation is a set of appearance changing augmentations consisting random grayscaling, random color jitter and gaussian blurring. Finally, the default augmentations is a combination of both ventral and dorsal augmentations.

\textbf{Datasets:} We pretrain the Hyper-SimCLR model on the unlabeled split of the STL-10 dataset consisting of 100000 images with each image is of size $96 \times 96$. We resize these images to $224 \times 224$ to match the image size of the downstream tasks. To evaluate the Hyper-SimCLR model on downstream tasks, we perform experiments on 300W landmark detection and CIFAR-10 image classification. On 300W use only the indoor sets where we use 40\% of the images to form a test set. For CIFAR-10, 80 \% of the labeled data available is used for learning invariances.

\textbf{Training details:} During the pre-trainig stage, the hypernetwork is trained for 200 epochs with the Adam optimizer with a learning rate of $3e^{-4}$, and weight decay coefficient of $1e^{-4}$ along with a cosine annealing learning rate scheduler. During the fine-tuning stage, the invariance hyper-parameters and corresponding task-specific prediction heads are trained for 100 epochs with the Adam optimizer using a learning rate of $3e^{-4}$ with a multi-step learning rate scheduler that decays the learning rate of each parameter group by $\gamma = 0.1$ after every 10 epochs. In this stage, we apply weight decay coefficient of $8e^{-4}$ only to the parameters of the linear readout layers.
\end{document}